# A Bilayer Segmentation-Recombination Network for Accurate Segmentation of Overlapping *C. elegans*

Mengqian Ding[a,b], Jun Liu[a,b,*], Yang Luo[a,b], and Jinshan Tang[c*]

[a]*Wuhan University of Science and Technology, Wuhan, 430081, China*

[b]*Hubei Province Key Laboratory of Intelligent Information Processing and Real-time Industrial System, Wuhan, China*

[c]*AI in Medicine Lab, College of Public Health, George Mason University, Fairfax, VA 22312, USA*

**Abstract**

Caenorhabditis elegans (*C. elegans*) is an excellent model organism because of its short lifespan and high degree of homology with human genes, and it has been widely used in a variety of human health and disease models. However, the segmentation of *C. elegans* remains challenging due to the following reasons: 1) the activity trajectory of *C. elegans* is uncontrollable, and multiple nematodes often overlap, resulting in blurred boundaries of *C. elegans*. This makes it impossible to clearly study the life trajectory of a certain nematode; and 2) in the microscope images of overlapping *C. elegans*, the translucent tissues at the edges obscure each other, leading to inaccurate boundary segmentation. To solve these problems, a Bilayer Segmentation-Recombination Network (BR-Net) for the segmentation of *C. elegans* instances is proposed. The network consists of three parts: A Coarse Mask Segmentation Module (CMSM), a Bilayer Segmentation Module (BSM), and a Semantic Consistency Recombination Module (SCRM). The CMSM is used to extract the coarse mask, and we introduce a Unified Attention Module (UAM) in CMSM to make CMSM better aware of nematode instances. The Bilayer Segmentation Module (BSM) segments the aggregated *C. elegans* into overlapping and non-overlapping regions. This is followed by integration by the SCRM, where semantic consistency regularization is introduced to segment nematode instances more accurately. Finally, the effectiveness of the method is verified on the *C. elegans* dataset. The experimental results show that BR-Net exhibits good competitiveness and outperforms other recently proposed instance segmentation methods in processing *C. elegans* occlusion images.

[*]Corresponding author

*Email address:* ljwhcn@qq.com (Jun Liu), jtang25@gmu.edu (Jinshan Tang)

---

## 1. Introduction

*C. elegans* is an excellent model organism for studying aging due to its ease of maintenance in the laboratory and its transparent body [1], which is easy to dissect and observe. Additionally, it shares genetic homology with humans (60-80%), has a complete genome sequence, exhibits conserved biomolecular responses, and demonstrates high fecundity (250 eggs per worm in just a few days) [2]. In addition, *C. elegans* has advantages such as a short lifespan of approximately three weeks and a small size, which reduces experimental costs and facilitates high-throughput screening experiments. This makes it an ideal candidate for screening anti-aging drugs [3]. Furthermore, nematode experiments do not raise ethical concerns. These benefits have contributed to numerous groundbreaking discoveries in the field of aging research [4].

When we use *C. elegans* for aging research, segmentation of individual worms is essential to accurately analyze specific anatomical regions, such as the pharynx, intestine, and reproductive system. This process enables the quantification of molecular and cellular changes, including protein aggregation, lipid accumulation, and mitochondrial dysfunction—key indicators of aging. Additionally, segmentation facilitates the tracking of spatially localized responses to genetic manipulations or pharmacological treatments, providing a detailed understanding of aging processes and the efficacy of potential anti-aging interventions. In the past, segmentation was often performed manually, which was both labor-intensive and time-consuming, limiting the scalability and reproducibility of experiments. Senescence studies require precise tracking and measurement of features such as body length, movement trajectories, and behavioral patterns. The development of instance segmentation techniques has revolutionized this field, enabling researchers to efficiently and accurately isolate the body of a nematode from the background. This advancement not only reduces labor but also enhances the precision of measurements, offering a clearer understanding of the biological changes that occur during aging. In recent years, deep learning has achieved remarkable success in image processing, prompting many researchers to apply these techniques in biological research. As a result, the study of *Caenorhabditis elegans* has evolved beyond traditional manual observation, with deep learning methods enabling more efficient and accurate analysis.

One challenge in segmenting individual worms arises when nematodes overlap or occlude one another (e.g., as shown in Figure 1b). In such cases, the pixels in the image may represent a mixture of multiple nematodes, making it difficult for the segmentation algorithm to accurately distinguish between them. This can lead to errors in instance segmentation, where parts of one nematode are incorrectly assigned to another. Overlapping can also distort the shape of the nematodes, as their boundaries may appear blurred in the image (e.g., as shown in Figure 1a). These issues hinder the algorithm's ability to accurately capture the true shape of each nematode, complicating the task of reliably segmenting an individual *C. elegans* under the microscope.

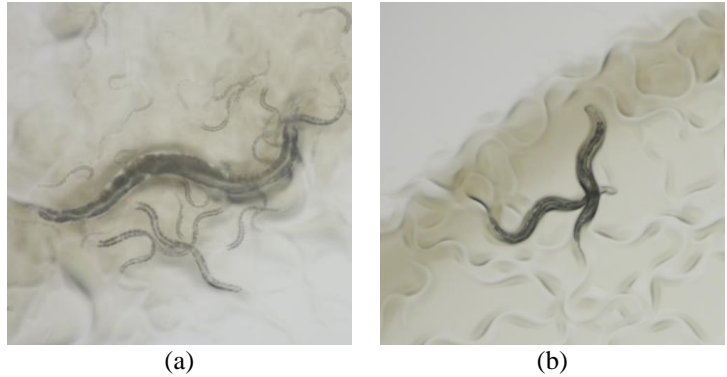

(a)　　　　　　　　　　(b)

Figure 1: Overlapping morphology of *C. elegans*.

Although there are several deep learning methods available for *C. elegans* image segmentation, accurate segmentation of occluded *C. elegans* remains an area with limited research. The initial instance segmentation method often results in inaccurate segmentation boundaries and an inability to accurately segment multiple aggregated *C. elegans*. This, in turn, makes it impossible to study the trajectory of the life activities of a particular nematode with accuracy. To solve the above problems, this paper proposes a new network for de-overlapping instance segmentation of *C. elegans* (BR-Net). The major contributions of this work are summarized as follows:

- To significantly enhance the network's ability to suppress irrelevant or noisy information, this paper introduces a Unified Attention Module (UAM) in the feature extraction process. The module specifically targets the removal of distractions in *C. elegans* images, such as food residues or excrement in the background, which often hinder accurate analysis. By focusing attention on the relevant features, UAM improves the clarity and quality of the extracted data, leading to better model performance.

- We propose a novel Bilayer Segmentation Module (BSM), designed to decompose overlapping *C. elegans* images into both overlapping and non-overlapping regions. This decomposition enables the model to handle complex image scenarios with more precision, distinguishing between regions where the *C. elegans* overlap and those that do not. This approach not only addresses segmentation challenges but also enhances the ability to isolate and identify individual elements with greater accuracy.

- Furthermore, we introduce the Semantic Consistency Recombination Module (SCRM), which utilizes semantic consistency regularization to improve the model's understanding of occlusion and overlap situations in *C. elegans* images. By enforcing semantic consistency across the different image regions, SCRM helps the model better interpret overlapping instances and occlusions, thereby significantly improving the accuracy of segmentation results and enhancing overall performance in challenging conditions.

## 2. Related work

### 2.1. Segmentation of Caenorhabditis elegans

In recent years, deep learning has achieved remarkable results in the field of image segmentation and classification [5-8]. Deep learning methods use multilayer networks to perform convolution and pooling operations directly on images, extracting key features. This approach offers significant advantages in image segmentation [9-12], including for segmenting *C. elegans.* For example, in 2020, Zeng et al. [13] enhanced Mask R-CNN with multilevel feature pooling and fusion branches to predict nematode contours. Wang et al. [14] developed a method to detect, segment, and locate pixel coordinates of *C. elegans* internal structures. In 2022, Xu et al. [15] proposed a method based on instance segmentation [16] to deal with the complexity of nematode cell shapes.

However, these basic segmentation methods described above have limitations: when performing image segmentation or instance segmentation on *C. elegans*, most studies tend to ignore the occluded regions. This may lead to errors in predicting the behavior and lifespan of *C. elegans* nematodes. It is important to consider all regions of *C. elegans* to obtain accurate results.

### 2.2. Amodal Instance Segmentation

To solve the above problem, Amodal instance segmentation has been proposed in the field of computer vision in recent years. The concept of 'amodal perception', originally proposed by psychologists, refers to the ability to infer the physical structure of an object even when parts of it are invisible. In 2016, Li and colleagues proposed a method to address the amodal instance segmentation problem [17]. The method involves randomly cropping the image to obtain patches and superimposing instances of other objects on the cropped image. The position and size of these superimposed objects are adjusted to ensure moderate overlap. The first step is to find the minimum bounding box of the visible portion for each superimposed object. Then, these boxes are randomly dithered to simulate the localization noise during testing. Finally, the pixels belonging to objects, background, or other objects in the image are labeled separately by generating target segmentation masks. Besides the work in [17], there have proposed a series work for Amodal instance segmentation [18-21]. In 2019, Follman et al. introduced Occlusion R-CNN (ORCNN) [18], extending Mask R-CNN with heads for amodal and occlusion masks. In 2021, Ke et al. proposed BCNet [19], a network with overlapping layers to detect occluding and partially occluded objects. Xiao et al. [20] also introduced a framework that refines non-visible masks by focusing on visible regions and using a shape prior for improved accuracy. In 2002, Sun et al. [21] replaced the classifier with a Bayesian model to capture learned features, trained on non-occluded images, and extended it for object segmentation, including occlusion.

Based on the above, amodal instance segmentation is well-suited for *C. elegans* segmentation due to its ability to manage occlusions and overlapping structures. *C. elegans* often appear partially hidden or overlapping in microscopic images, but amodal segmentation can predict both visible and occluded parts, enabling accurate segmentation even when portions of the worm are obscured. This enhances tracking, quantification, and analysis of the worm's anatomy and behavior, making it highly effective for biological studies. Our proposed method will leverage this approach.

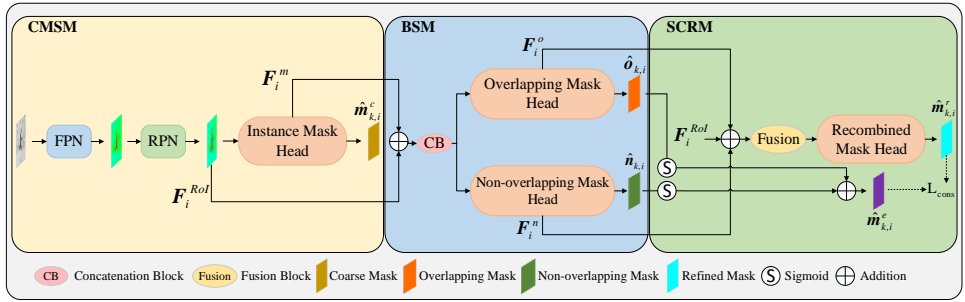

Figure 2: BR-Net network structure. Inspired by the study in literature [18], this paper utilises the Coarse Mask Segmentation Module (CMSM) in BR-Net to obtain the coarse mask. The Bilayer Segmentation Module (BSM) is then employed, taking the instance features as inputs, to predict overlapping and non-overlapping regions of the *C. elegans*. After decomposing the aggregated nematodes based on positional relationships, the Semantic Consistency Recombination Module (SCRM) generates recombined refinement masks using instance features from the BSM and RoIAlign layers and performs semantic consistency regularization.

## 3. Methods

The structure of the *C. elegans* de-overlapping deep neural network proposed in this paper is shown in Figure 2. The net is composed of three modules: Coarse Mask Segmentation Module (CMSM), Bilayer Segmentation Module (BSM), and Semantic

Consistency Recombination Module (SCRM). We will describe them in detail in the following subsections.

*3.1. Problem formulation*

A *C. elegans* dataset D consisting of K images and their corresponding annotations. Each image contains annotations of bounding box $B_k = \{b_{k,i}\}_{i=1}^{A_k}$ and instance mask $M_k = \{m_{k,i}\}_{i=1}^{A_k}$, where $A_k$ represents the number of instances in the k-th image. For each nematode group, we decompose the instance mask into overlapping regions $O_k = \{o_{k,i}\}_{i=1}^{A_k}$ and non-overlapping regions $N_k = \{n_{k,i}\}_{i=1}^{A_k}$ through logical operations based on the positional relationship between nematodes.

*3.2. Coarse Mask Segmentation Module (CMSM)*

In the Coarse Mask Segmentation Module, Mask R-CNN is adopted as the basic framework. Previous research has demonstrated the competitive performance of Mask R-CNN in instance segmentation [18]. Mask R-CNN comprises two stages: the first stage performs feature extraction using the Feature Pyramid Network (FPN) and generates candidate object bounding boxes using the Region Proposal Network (RPN). The second stage generates a region of interest (RoI) features $F_i^{RoI}$ through the RoIAlign layer, which predicts the bounding box $\hat{b}_{k,i}$ from the detection head while predicting the semantic mask $\hat{m}_{k,i}^c$ from the instance mask head. As overlapping regions may limit perceptual ability, $\hat{m}_{k,i}^c$ may contain fuzzy boundaries. In this paper, $\hat{m}_{k,i}^c$ is denoted as a coarse mask, providing information for sub-region decomposition in the BSM and suppressing interference from the background. The multitask loss L for coarse mask segmentation follows the standard loss function described in [12]. The calculation of the multi-task loss for coarse mask segmentation $L_{coarse}$ is as follows:

$$L_{coarse} = L_{cls} + L_{reg} + L_{cmask} \quad (1)$$

where $L_{cls}$ represents the cross-entropy (CE) loss for classification, $L_{reg}$ represents the Smooth L1 Loss for bounding box regression, and $L_{cmask}$ represents the pixel-wise cross-entropy (CE) loss for segmentation. The calculation formula of $L_{cls}$, $L_{reg}$ and $L_{cmask}$ are as follows:

$$L_{cls} = -\frac{1}{K}\sum_{k=1}^{K}\sum_{i=1}^{A_k} y_{i,c} \cdot log(p_{i,c}) \quad (2)$$

where K represents the number of images in the *C. elegans* dataset D. $A_k$ represents the number of instances in the k-th image. $y_{i,c}$ is the true label of the i-th sample, and its value is 0 or 1. When the i-th sample category is c, $y_{i,c} = 1$, otherwise $y_{i,c} = 0$. $p_{i,c}$ is the probability of predicting the i-th sample as category c.

$$L_{reg} = smooth_{L_1}(x) = \begin{cases} 0.5x^2, |x| < 1 \\ |x| - 0.5, |x| \geq 1 \end{cases} \quad (3)$$

Since L1 loss is more robust than L2 loss, it is less sensitive to outliers. When the regression target is unbounded, it is easy to explode the gradient using L2 loss, so equation (3) uses L1 loss to avoid this sensitivity.

$$L_{cmask} = -(y_i \cdot log(p_i) + (1 - y_i) \cdot log(1 - p_i)) \quad (4)$$

where $y_i$ is the true label (0 or 1), $p_i$ is the probability of predicting this pixel as category 1, and $1 - p_i$ is the probability of predicting this pixel as category 0.

*3.3. Bilayer Segmentation Module (BSM)*

BSM is utilized to segment the overlapping regions of *C. elegans*. BSM comprises overlapping and non-overlapping mask heads with identical architecture. The semantic features in the instance mask header before coarse mask prediction are denoted by $\boldsymbol{F}_i^m$. Pass $\boldsymbol{F}_i^{RoI}$ and $\boldsymbol{F}_i^m$ through a connection block as the input of the overlapping mask head and the non-overlapping mask head, so as to predict the overlapping area $\widehat{\boldsymbol{o}}_{k,i}$ and the non-overlapping area $\widehat{\boldsymbol{n}}_{k,i}$ in the instance. Among them, each mask head consists of 4 convolution layers and 1 deconvolution layer. The convolution layer is used to generate 14×14×256 features, and the deconvolution layer is used to obtain a resolution of 28×28×1 semantic mask. Pixel-wise CE loss $L_{ce}$ is added to both heads as an explicit constraint for decomposition. $L_{dec}$ is the loss function used in BSM, and its calculation formula is as follows:

$$L_{dec} = \frac{1}{K}\sum_{k=1}^{K}\frac{1}{A_k}\sum_{i=1}^{A_k}\left(L_{cmask}(\widehat{\boldsymbol{o}}_{k,i}, \boldsymbol{o}_{k,i}) + L_{cmask}(\widehat{\boldsymbol{n}}_{k,i}, \boldsymbol{n}_{k,i})\right) \quad (5)$$

*3.4. Semantic Consistency Recombination Module (SCRM)*

To improve the perception of overlapping instances, SCRM incorporates contextual information to better perceive the overall instance. The features before the last layer in the Overlapping Mask Head and Non-overlapping Mask Head are represented by $\boldsymbol{F}_i^o$ and $\boldsymbol{F}_i^n$. The semantic features between overlapping and non-overlapping areas often contain information about the mutual influence and interaction between objects in complex scenes, which is considered to be the residual information of complex areas. $\boldsymbol{F}_i^o$ and $\boldsymbol{F}_i^n$ pass through the fusion block together with $\boldsymbol{F}_i^{RoI}$, and then are input into the Recombined Mask Head. This paper introduces fusion blocks to SCRM and reuses RoI features to predict instances. This helps SCRM to utilize contextual information of overlapping instances, thereby improving its perception ability and optimizing the thinning mask $\widehat{\boldsymbol{m}}_{k,i}^r$ from SCRM through segmentation loss $L_{rmask}$. The calculation of the $L_{rmask}$ is as follows:

$$L_{rmask} = \frac{1}{K} \sum_{k=1}^{K} \frac{1}{A_k} \sum_{i=1}^{A_k} \left( L_{cmask}(\widehat{\boldsymbol{m}}_{k,i}^r, \boldsymbol{m}_{k,i}) \right) \tag{6}$$

Due to occlusion and overlap, *C. elegans* images often have unclear boundaries for each instance, making segmentation results blurred. Instance segmentation models of the traditional kind may face challenges in accurately segmenting each instance of *C. elegans*. To address this issue, the paper introduces semantic consistency regularization $L_{cons}$, which relates the recombination mask $\widehat{\boldsymbol{m}}_{k,i}^r$ to the merged predictions of $\widehat{\boldsymbol{o}}_{k,i}$ and $\widehat{\boldsymbol{n}}_{k,i}$. The semantic consistency regularization enhances the model's semantic comprehension of occlusion and overlapping scenarios and ensures that instances are properly separated while maintaining semantic consistency. As a result, the model can better identify the boundaries and shapes of nematodes in *C. elegans* images, leading to improved segmentation accuracy. The calculation of the $L_{cons}$ is as follows:

$$L_{cons} = \frac{1}{K} \sum_{k=1}^{K} \frac{1}{A_k} \sum_{i=1}^{A_k} L_{cmask} \left( \widehat{\boldsymbol{m}}_{k,i}^r, \boldsymbol{F}_{merge}(\widehat{\boldsymbol{o}}_{k,i}, \widehat{\boldsymbol{n}}_{k,i}) \right) \tag{7}$$

The formula for merging the overlapping $\widehat{\boldsymbol{o}}_{k,i}$ and non-overlapping $\widehat{\boldsymbol{n}}_{k,i}$ areas is represented by $\boldsymbol{F}_{merge}(\cdot)$, which is equivalent to the XOR$(\cdot)$ operation. The sub-region masks are normalized using the Sigmoid function. The XOR of the two mask logics is then computed to merge them and suppress redundant prediction pixels.

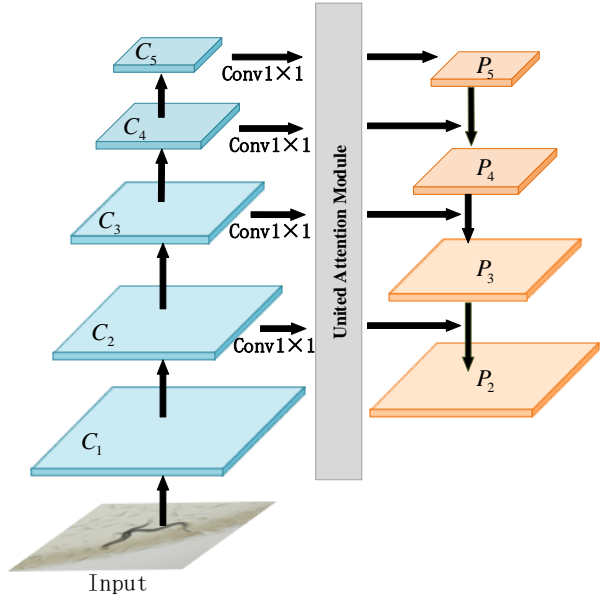

Figure 3: Feature Extraction Network of Mask R-CNN Model Improved by United Attention Module.

*3.5. United Attention Module (UAM)*

This article introduces a United Attention Module (UAM) in CMSM. The UAM is added respectively after the 1×1 convolutional layers of $C_1$, $C_2$, $C_3$, and $C_4$ to improve the feature fusion stage of the model feature extraction network. The UAM enhances the channel and spatial focusing capabilities of CMSM on *C. elegans* populations, as demonstrated in Figure 3. The UAM comprises of three key components: convolutional layers with varying kernel sizes, a channel attention module, and a spatial attention module, as illustrated in Figure 4. The input feature map undergoes three parallel convolutional layers to generate three feature maps with distinct receptive fields. These convolutional layers consist of a 5×5 convolution, a 3×3 dilated convolution with a dilation rate of 3, and a 3×3 convolution. The feature maps obtained from the three convolutional layers can be represented as:

$$F_5^C = F_{input} \times MT_{5\times5} \tag{8}$$

$$F_D^C = F_{input} \times MT_{3\times3}^D \tag{9}$$

$$F_3^C = F_{input} \times MT_{3\times3} \tag{10}$$

where $F_3^C$ and $F_5^C$ represent the feature maps captured by the 3×3 convolution and 5×5 convolution layers respectively. $F_D^C$ represents the feature map captured by dilated convolution. $F_{input}$ represents the input feature map, $MT_{3\times3}$ and $MT_{5\times5}$ represent the matrices of 3×3 convolution and 5×5 convolution respectively. $MT_{3\times3}^D$ represents the matrix of dilated convolution.

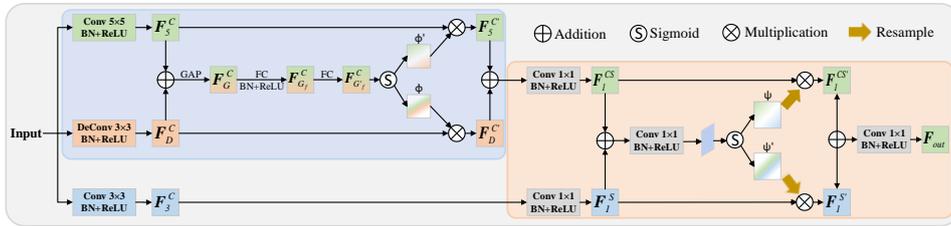

Figure 4: United Attention Module. The input image first passes through the channel attention module, and then through the spatial attention module.

*3.5.1. Channel attention module*

The channel attention module helps enhance the expressive power of channels with rich semantic information. To capture useful objective features from different receptive fields, this paper introduces a channel attention module that guides the network to learn more robust feature representations. The blue section in Figure 4 depicts the channel attention module. The channel attention module enables the network to focus on the most relevant channels for the task, enhancing the model's sensitivity to key features. This results in the selection of more representative features from the channel dimension. Specifically, the module compresses the combined feature map of $F_5^C$ and $F_D^C$ into a new feature map $F_G^C$ of size 1×1 using global average pooling (GAP). The new feature map can be expressed as:

$$F_G^C = \text{GAP}(F_5^C \oplus F_D^C) \quad (11)$$

The symbol ⊕ denotes element-wise addition. Next, the feature map $F_G^C$ is fed into the fully connected layer, followed by the batch normalization (BN) layer and the activation function ReLU layer, resulting in a new feature map $F_{G_f}^C$. The expression for $F_{G_f}^C$ is as follows:

$$F_{G_f}^C = \text{ReLU}(\text{B}(MT_{FC} \cdot F_G^C)) \quad (12)$$

The matrix of the fully connected layer is represented by $MT_{FC}$. Batch normalization (BN) and ReLU activation operations are represented by B(·) and ReLU(·) respectively. To obtain a new feature map $F_{G_f'}^C$, perform the full connection operation on the feature map $F_{G_f}^C$ again, expressed as follows:

$$F_{G_f'}^C = MT_{FC} \cdot F_{G_f}^C \quad (13)$$

Finally, apply the sigmoid activation function to the feature map $F_{G_f'}^C$ in order to obtain the channel attention map $\varphi$：

$$\varphi = \sigma\left(F_{G_f'}^C\right) \quad (14)$$

where $\varphi \in [0,1]$ and $\varphi' \in [0,1]$ represent the channel attention maps of $F_D^C$ and $F_5^C$ respectively, where $\varphi' = 1 - \varphi$. Each value of $\varphi / \varphi'$ represents the importance of the channel information at the corresponding pixel in $F_D^C / F_5^C$. This article uses the channel attention map $\varphi$ to calibrate the feature map $F_D^C$，and uses the channel attention map $\varphi'$ to calibrate the feature map $F_5^C$. The feature map calibrated by the channel attention map can be expressed as:

$$F_D^{C'} = \varphi \otimes F_D^C \quad (15)$$

$$F_5^{C'} = \varphi' \otimes F_5^C \quad (16)$$

The feature maps $F_D^{C'}$ and $F_5^{C'}$ are then integrated and used as input for the spatial attention module.

*3.5.2. Spatial attention module*

Previous research has concluded that channel attention aims to establish the relationship between different channels, allowing the network to prioritize important feature channels and suppress unimportant ones. Meanwhile, spatial attention aims to learn the characteristics of different positions in the feature map, enabling the network to focus on important spatial regions [22]. To enhance the robustness of network feature representation, this article introduces a spatial attention module following the channel attention module (shown in green in the Figure 4). To refine the target's position information, a 1x1 convolution operation is performed on the input feature map. The resulting feature map is defined as:

$$F_1^{CS} = \left(F_5^{C'} \oplus F_D^{C'}\right) \cdot MT_{1\times1} \tag{17}$$

$$F_1^S = F_3^C \cdot MT_{1\times1} \tag{18}$$

Subsequently, the fused feature maps of $F_1^{CS}$ and $F_1^S$ are subjected to ReLU activation, followed by a 1×1 convolution operation and sigmoid activation, to obtain the spatial attention map $\psi$. This can be expressed using the following formula:

$$\psi = \sigma(\text{ReLU}(F_1^{CS} \oplus F_1^S) \cdot MT_{1\times1}) \tag{19}$$

where $\psi \in [0,1]$ and $\psi' \in [0,1]$ represent the spatial attention maps of $F_1^{CS}$ and $F_1^S$ respectively, where $\psi' = 1 - \psi$. Each value of $\psi / \psi'$ represents the importance of the spatial information at the corresponding pixel in $F_1^{CS} / F_1^S$. To calibrate $F_1^{CS}$ and $F_1^S$, $\psi$ and $\psi'$ are resampled to obtain spatial attention maps with the same number of channels as $F_1^{CS}$ and $F_1^S$. The calibrated feature maps are denoted as $F_1^{CS'}$ and $F_1^{S'}$ respectively. Finally, $F_1^{CS'}$ and $F_1^{S'}$ are fused and convolved to obtain the output $F_{out}$ of UAM. The calculation of the $F_{out}$ is as follows:

$$F_{out} = \left(F_1^{CS'} \oplus F_1^{S'}\right) \cdot MT_{1\times1} \tag{20}$$

### 3.6. Loss function

The instance segmentation framework proposed in this paper for *C. elegans* is trained in a fully supervised manner. The target loss L is defined as follows:

$$L = L_{\text{coarse}} + \lambda_{\text{dec}} L_{\text{dec}} + \lambda_{\text{rmask}} L_{\text{rmask}} + \lambda_{\text{cons}} L_{\text{cons}} \tag{21}$$

The loss for RoI extraction and coarse mask prediction is represented by $L_{\text{coarse}}$, while $L_{\text{dec}}$ represents the decomposition loss for overlapping and non-overlapping region segmentation. $L_{\text{rmask}}$ is the segmentation loss for refined masks, and $L_{\text{cons}}$ supervises the semantic consistency between the overall instance and sub-regions. $\lambda_{\text{dec}}$, $\lambda_{\text{rmask}}$ and $\lambda_{\text{cons}}$ are trade-off parameters that control the importance of each component.

## 4. Experiments and Analysis

### 4.1. Dataset

While *C. elegans* is a well-studied organism in biology, there are limited publicly available and fully annotated image datasets for this species. To evaluate the effectiveness of the method presented in this article, we used live videos of *C. elegans* captured under a microscope by the Sino-French Joint Laboratory of the School of Life Science and Technology at Huazhong University of Science and Technology as depicted in Figure 5 [24]. To train the proposed instance segmentation algorithm, for each video sequence, one frame was extracted every 3 seconds and saved. Furthermore, image processing methods, including random rotation, random cropping, and random horizontal flipping, were used to process the images and mask in the training set. Finally, we obtained two *C. elegans* datasets, C. Data-1 and C. Data-2, collected from different time periods. Among these, C. Data-1 contains 453 images, which mainly recorded the early life stage of *C. elegans*. These images have less background noise, which is conducive to instance segmentation. C.Data-2 contains 639 images, which recorded the middle and late life stages of C. elegans. These images have large background noise, including food residues, excrement, and other *C. elegans* corpses, which increases the difficulty of segmentation. In total, the dataset comprises 1,092 annotated images, with manually generated bounding boxes and mask labels created using the annotation tool LabelMe. The dataset was split into a training set comprising 880 images and a test set containing 212 images.

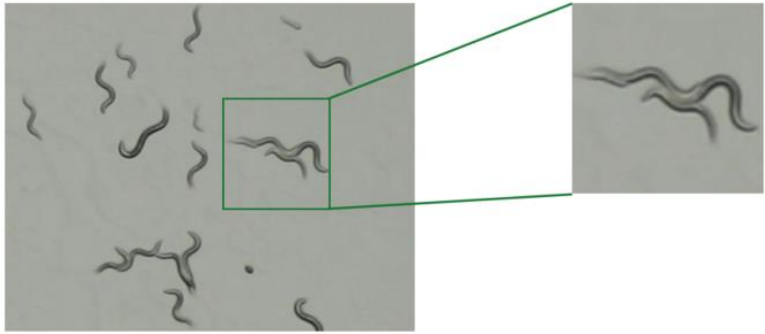

Figure 5: *C. elegans* dataset sample.

*4.2. Evaluation Measurement*

To evaluate the performance of the proposed BR-Net, we used four widely adopted instance segmentation metrics: average precision (AP), average precision at IoU thresholds of 0.5 ($AP_{50}$) and 0.75 ($AP_{75}$), and mean intersection over union (mIoU). Precision measures the proportion of correctly identified positive instances, while recall quantifies the model's ability to detect true instances. AP, the area under the precision-recall (P-R) curve, captures average accuracy across different recall levels. IoU (Intersection over Union) quantifies the overlap between predicted and ground truth boundaries, with $AP_{50}$ and $AP_{75}$ calculated at IoU thresholds of 0.5 and 0.75, respectively. mIoU, the average IoU across all categories, is particularly important for overlapping segmentation tasks as it evaluates both the accuracy of individual instance segmentation and the overall performance in handling multiple overlapping instances. The calculation formulas for each evaluation indicator are as follows:

$$Precision = \frac{TP}{TP+FP} \tag{22}$$

$$Recall = \frac{TP}{TP+FN} \tag{23}$$

$$AP = \int_0^1 PR(r)\,dr \tag{24}$$

$$IoU = \frac{TP}{TP+FN+FP} \tag{25}$$

$$mIoU = \frac{1}{k+1}\sum_{i=0}^{k}\frac{TP}{TP+FN+FP} \tag{26}$$

where *PR(r)* is the precision at recall *r*. TP (True Positives) refers to positive category samples that are correctly predicted as positive categories; TN (True Negatives) refers to negative category samples that are correctly predicted as negative categories; FP (False Positives) refers to negative category samples that are incorrectly predicted as positive categories; FN (False Negatives) refers to positive category samples that are incorrectly predicted as negative categories.

*4.3. Parameter setting and training*

The article implements the BR-Net method using PyTorch 1.11.0 and trains it on a server with two NVIDIA RTX3090 GPU 24 GB graphics cards. The experiment uses Mask R-CNN as the baseline model and the FPN network based on ResNet-50. The training process sets the batch size to 8 and the initial learning rate to 0.01, which gradually decreases to 0.001. Additionally, linear warmup is added in the first 20 iterations. The network is trained for 200 iterations using Adam as the optimizer in this paper. A comparative experiment is conducted between this method and other instance segmentation methods, followed by an ablation experiment on each component of this method.

*4.4. Result and Analysis*

*4.4.1. Comparative Experiments on Various Instance Segmentation Methods*

We conducted comparative experiments on two datasets, C. Data-1 and C. Data-2, to evaluate the performance of classic segmentation methods [12]-[14][23], amodal instance segmentation methods [17]-[20], and worm-specific approaches [25]. The results indicate that worm-specific methods generally outperform classic approaches in segmentation accuracy, although they are slightly less efficient in terms of runtime. Notably, methods designed to address de-overlapping significantly enhance segmentation performance. As shown in Table 1, BR-Net achieves the highest scores across all key evaluation metrics, including AP, $AP_{50}$, $AP_{75}$, and mIoU, demonstrating its superior segmentation capabilities. While BR-Net is marginally slower than classic methods like Mask R-CNN in runtime, it offers an excellent balance between efficiency and accuracy. For the C. Data-2 dataset, the increased complexity of background interference results in reduced segmentation performance for all methods compared to C. Data-1. Nevertheless, BR-Net maintains its leading performance across all metrics, showcasing its exceptional robustness. These results highlight BR-Net's strong adaptability to complex environments and its potential for practical applications.

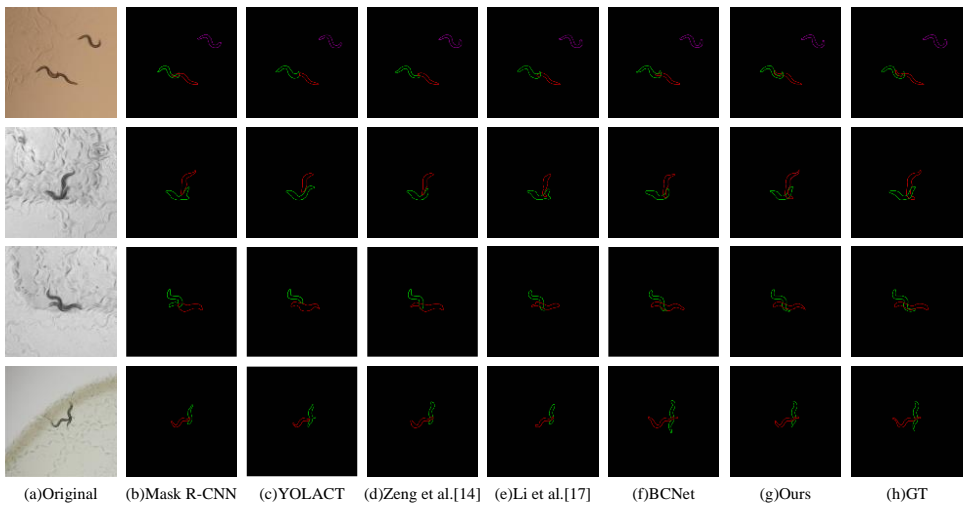

(a)Original   (b)Mask R-CNN   (c)YOLACT   (d)Zeng et al.[14]   (e)Li et al.[17]   (f)BCNet   (g)Ours   (h)GT

Figure 6: Segmentation results of *C. elegans* in different methods. Figure 6 selects four sets of *C. elegans* images with occlusion phenomenon and compares this paper's method with the remaining five methods. The comparison includes traditional methods, previously proposed methods for segmenting *C. elegans* [14], and the amodal instance segmentation method [17][19].

Since our method aims to accurately segment smaller instances, it achieves relatively high scores across all metrics. Compared to BCNet, there is an increase of 7.8 percentage points in $AP_{50}$. This improvement is attributed to our method's preference for segmenting smaller objects. Figure 6 presents the visual results of comparative experiments between BR-Net and other methods on this paper's dataset. Figure 6 shows that some methods produce under-segmentation and over-segmentation in the segmentation results of overlapping *C. elegans* clusters. Previous networks, such as Mask R-CNN (Figure 6(b)) and YOLACT (Figure 6(c)), struggle to capture the relationship between pixels in overlapping and non-overlapping sub-regions, leading to blurry segmentations. In contrast, our BR-Net effectively distinguishes different instance boundaries and accurately perceives the completeness of *C. elegans*.

Table 1: Performance comparison of BR-Net and other methods for segmentation of *C. elegans*. Bolded numbers indicate optimal results.

| Method | Dataset | Evaluation index | | | | |
| --- | --- | --- | --- | --- | --- | --- |
| | | AP↑ | AP50↑ | AP75↑ | mIoU↑ | Test Cost(s)↓ |
| Mask R-CNN [12] | C.Data-1 | 0.461 | 0.627 | 0.529 | 0.495 | 1.83 |
| YOLACT [13] | | 0.453 | 0.631 | 0.446 | 0.517 | 1.75 |
| Mask Scoring R-CNN [23] | | 0.507 | 0.689 | 0.513 | 0.576 | **1.69** |
| Zeng et al. [14] | | 0.565 | 0.726 | 0.576 | 0.622 | 2.72 |
| Li et al. [17] | | 0.597 | 0.751 | 0.608 | 0.651 | 3.26 |
| Occlusion R-CNN [18] | | 0.648 | 0.784 | 0.686 | 0.693 | 2.80 |
| BCNet [19] | | 0.661 | 0.803 | 0.675 | 0.714 | 2.91 |
| Xiao et al. [20] | | 0.683 | 0.812 | 0.691 | 0.747 | 3.39 |
| WormSwin [25] | | 0.721 | 0.850 | 0.737 | 0.765 | 2.53 |
| BR-Net(ours) | | **0.739** | **0.881** | **0.760** | **0.803** | 2.07 |
| Mask R-CNN [12] | C.Data-2 | 0.437 | 0.559 | 0.463 | 0.446 | 1.95 |
| YOLACT [13] | | 0.426 | 0.568 | 0.417 | 0.461 | **1.81** |
| Mask Scoring R-CNN [23] | | 0.459 | 0.582 | 0.476 | 0.493 | 1.88 |
| Zeng et al. [14] | | 0.472 | 0.615 | 0.513 | 0.525 | 2.94 |
| Li et al. [17] | | 0.521 | 0.661 | 0.536 | 0.562 | 3.51 |
| Occlusion R-CNN [18] | | 0.557 | 0.693 | 0.588 | 0.633 | 2.97 |
| BCNet [19] | | 0.603 | 0.729 | 0.622 | 0.680 | 3.08 |
| Xiao et al. [20] | | 0.644 | 0.753 | 0.673 | 0.691 | 3.42 |
| WormSwin [25] | | 0.682 | 0.802 | 0.705 | 0.736 | 2.69 |
| BR-Net(ours) | | **0.712** | **0.858** | **0.754** | **0.791** | 2.26 |

*4.4.2. Ablation experiment*

This paper presents the results of two ablation experiments. The first experiment aimed to investigate the effect of different modules in the proposed BR-Net on various evaluation indicators. The comparison results are presented in Table 2. The introduction of BSM enabled successful segmentation of clustered *C. elegans* instances into overlapping and non-overlapping areas. Table 2 shows that the $AP_{50}$ in the C. Data-1 data set improved by 15%. However, incorporating BSM without considering structural and morphological information may mislead the model. To address this issue, BR-Net adopts a semantic consistency recombination strategy and adds SCRM after BSM. SCRM improved the $AP_{50}$ on the C. Data-2 data set by 5.3 percentage points by enhancing the model's perception of overlapping areas while retaining morphological information.

Table 2: Ablation experiments for each module in BR-Net.

| Method | Dataset | Evaluation index | | | |
|---|---|---|---|---|---|
| | | AP↑ | AP50↑ | AP75↑ | mIoU↑ |
| CMSM | C.Data-1 | 0.482 | 0.671 | 0.553 | 0.542 |
| CMSM+ BSM | | 0.668 | 0.821 | 0.719 | 0.764 |
| CMSM+ BSM+ SCRM | | **0.739** | **0.881** | **0.760** | **0.803** |
| CMSM | C.Data-2 | 0.451 | 0.602 | 0.485 | 0.527 |
| CMSM+ BSM | | 0.604 | 0.805 | 0.695 | 0.759 |
| CMSM+ BSM+ SCRM | | **0.712** | **0.858** | **0.754** | **0.791** |

The second experiment aims to investigate the impact of various components in the module on different evaluation indicators. We conducted a detailed comparison of different components in BR-Net using the *C. elegans* data set to demonstrate their effectiveness. Specifically, we evaluated the effectiveness of the following components: 1) $B_c$：Instance Mask Head used solely for coarse segmentation; 2) $B_o$：Overlapping Mask Head used for overlapping area segmentation; 3) $B_n$：The Non-overlapping Mask Head is used for segmenting non-overlapping areas. 4) UAM: United Attention Module for enhanced perception. 5) $L_{cons}$：Semantic consistency regularization between the recombination prediction result $\widehat{m}_{k,i}^r$ and the fusion sub-region $\widehat{o}_{k,i}$ and $\widehat{n}_{k,i}$. Table 3 shows that the addition of $B_o$ increases the $AP_{50}$ by 6.8 percentage points in the C.Data-2. On this basis, adding $B_n$ increases the $AP_{50}$ by 11.7 percentage points. Further adding UAM to $B_c$ to improve the channel and spatial focusing capabilities of CMSM on *C. elegans* populations, we observed a 6.1 percentage point increase in $AP_{50}$. Semantic consistency regularization aims to enhance the instance segmentation model's comprehension of image semantics, enabling it to handle scenes with occlusion, overlap, and blurred instance boundaries. Additionally, it improves the model's ability to reason about overlapping instances by teaching it the concept of reorganized instances in sub-regions. Table 3 demonstrates that the inclusion of semantic consistency regularization $L_{cons}$ results in a further 5.3 percentage point increase in $AP_{50}$.

Table 3: Ablation experiments of different components in each module.

| Method | | | | | Dataset | Evaluation index | | | |
|---|---|---|---|---|---|---|---|---|---|
| $B_c$ | $B_o$ | $B_n$ | UAM | $L_{cons}$ | | AP↑ | AP50↑ | AP75↑ | mIoU↑ |
| √ | | | | | | 0.461 | 0.627 | 0.529 | 0.495 |
| √ | √ | | | | | 0.549 | 0.664 | 0.575 | 0.619 |
| √ | √ | √ | | | C.Data-1 | 0.627 | 0.763 | 0.631 | 0.702 |
| √ | √ | √ | √ | | | 0.668 | 0.821 | 0.719 | 0.764 |
| √ | √ | √ | √ | √ | | **0.739** | **0.881** | **0.760** | **0.803** |
| √ | | | | | | 0.437 | 0.559 | 0.463 | 0.446 |
| √ | √ | | | | | 0.462 | 0.627 | 0.559 | 0.603 |
| √ | √ | √ | | | C.Data-2 | 0.547 | 0.744 | 0.612 | 0.698 |
| √ | √ | √ | √ | | | 0.604 | 0.805 | 0.695 | 0.759 |
| √ | √ | √ | √ | √ | | **0.712** | **0.858** | **0.754** | **0.791** |

## 5. Conclusion

Caenorhabditis elegans serves as a widely embraced miniature model organism in life science research, particularly within the domain of lifespan investigation. In this context, this paper introduces a Bilayer Segmentation-Recombination Network (BR-Net) designed to accurately segment overlapping *C. elegans* images, addressing the challenge of delineating such instances effectively. The network comprises three key components: the Coarse Mask Segmentation Module (CMSM), Bilayer Segmentation Module (BSM), and Semantic Consistency Recombination Module (SCRM). To enhance the perception of *C. elegans* instances, we propose the integration of a United Attention Module (UAM). Comparative experiments demonstrate that our method yields more precise segmentation results compared to existing models across both training and test sets. The BR-Net segmentation approach facilitates deeper analysis and comprehension of individual nematodes, thereby advancing life science research. Future endeavors will involve augmenting the dataset to accommodate the scarcity of *C. elegans* images, alongside refining the model's overall network structure, parameter configurations, and other aspects to continually enhance its performance and applicability.

[12] Guo, M. H., Lu, C. Z., Hou, Q., Liu, Z., Cheng, M. M., & Hu, S. M. (2022). Segnext: Rethinking convolutional attention design for semantic segmentation. *Advances in Neural Information Processing Systems*, *35*, 1140-1156..

[13] Zeng Z, Liu J., Microscopic image segmentation method of *C. elegans* based on deep learning[J]. Journal of Computer Applications, 2020, 40(5): 1453.

[14] Wang L, Kong S, Pincus Z, et al. Celeganser: Automated analysis of nematode morphology and age[C]//Proceedings of the IEEE/CVF Conference on Computer Vision and Pattern Recognition Workshops. 2020: 968-969.

[15] Xu R, Li Y, Wang C, et al. Instance segmentation of biological images using graph convolutional network[J]. Engineering Applications of Artificial Intelligence, 2022, 110: 104739.

[16] Bolya D, Zhou C, Xiao F, et al. Yolact: Real-time instance segmentation[C]//Proceedings of the IEEE/CVF international conference on computer vision. 2019: 9157-9166.

[17] Li K, Malik J. Amodal instance segmentation[C]//Computer Vision–ECCV 2016: 14th European Conference, Amsterdam, The Netherlands, October 11-14, 2016, Proceedings, Part II 14. Springer International Publishing, 2016: 677-693.

[18] Follmann P, König R, Härtinger P, et al. Learning to see the invisible: End-to-end trainable amodal instance segmentation[C]//2019 IEEE Winter Conference on Applications of Computer Vision (WACV). IEEE, 2019: 1328-1336.

[19] Ke L, Tai Y W, Tang C K. Deep occlusion-aware instance segmentation with overlapping bilayers[C]//Proceedings of the IEEE/CVF conference on computer vision and pattern recognition. 2021: 4019-4028.

[20] Xiao Y, Xu Y, Zhong Z, et al. Amodal segmentation based on visible region segmentation and shape prior[C]//Proceedings of the AAAI Conference on Artificial Intelligence. 2021, 35(4): 2995-3003.

[21] Sun Y, Kortylewski A, Yuille A. Amodal segmentation through out-of-task and out-of-distribution generalization with a Bayesian model[C]//Proceedings of the IEEE/CVF Conference on Computer Vision and Pattern Recognition. 2022: 1215-1224.

[22] Woo S, Park J, Lee J Y, et al. Cbam: Convolutional block attention module[C]//Proceedings of the European conference on computer vision (ECCV). 2018: 3-19.
19